\documentclass{ecai} 

\usepackage{latexsym}
\usepackage{amssymb}
\usepackage{amsmath}
\usepackage{amsthm}
\usepackage{booktabs}
\usepackage{enumitem}
\usepackage{graphicx}
\usepackage{color}
\usepackage{multirow}
\usepackage{xcolor}
\usepackage{amsmath}
\usepackage{float}
\usepackage{diagbox}

\newtheorem{theorem}{Theorem}
\newtheorem{lemma}[theorem]{Lemma}
\newtheorem{corollary}[theorem]{Corollary}

\newcommand{\BibTeX}{B\kern-.05em{\sc i\kern-.025em b}\kern-.08em\TeX}

\begin{document}


\begin{frontmatter}


\paperid{2269} 


\title{Partially Trained Graph Convolutional Networks Resist Oversmoothing}


\author[A,B]{\fnms{Dimitrios}~\snm{Kelesis}\thanks{Corresponding Author. Email: dkelesis@iit.demokritos.gr.}}
\author[A]{\fnms{Dimitris}~\snm{Fotakis}}
\author[B]{\fnms{Georgios}~\snm{Paliouras}}

\address[A]{School of Electrical Engineering and Computer Science, National Technical University of Athens, Greece}
\address[B]{National Center for Scientific Research ``Demokritos'', Greece}


\begin{abstract}
In this work we investigate an observation made by Kipf \& Welling, who suggested that untrained GCNs can generate meaningful node embeddings. In particular, we investigate the effect of training only a single layer of a GCN, while keeping the rest of the layers frozen. We propose a basis on which the effect of the untrained layers and their contribution to the generation of embeddings can be predicted. Moreover, we show that network width influences the dissimilarity of node embeddings produced after the initial node features pass through the untrained part of the model. Additionally, we establish a connection between partially trained GCNs and oversmoothing, showing that they are capable of reducing it. We verify our theoretical results experimentally and show the benefits of using deep networks that resist oversmoothing, in a ``cold start'' scenario, where there is a lack of feature information for unlabeled nodes.
\end{abstract}

\end{frontmatter}


\section{Introduction}
Graph Neural Networks (GNNs) have shown promising results in diverse graph analytics tasks, including node classification \cite{Zhang,graphsage}, link prediction \cite{link_pred1,link_pred2}, and graph classification \cite{chemistry}. Consequently, they play a key role in graph representation learning, employing message passing schemes to aggregate information from neighboring nodes and construct informative node representations. One of the most prominent GNN models is the Graph Convolutional Network (GCN) \cite{Kipf_gcn}, which generates node representations by computing the average of the representations (embeddings) of each node's immediate neighbors. \\
At the same time, the improved performance of deep Convolutional Neural Networks (CNNs) compared to their shallow counterparts, has led to several attempts towards building deep GNNs, in order to solve graph analytics tasks more effectively \cite{Li_et, JK_Nets}. Greater depth offers the potential for more accurate representational learning, by expanding the model's receptive field and integrating more information. Hence, there is a demand for the creation of new models capable of scaling efficiently to a significant number of layers. However, designing deep GNNs has proven to be challenging in practice, as increased depth often leads to performance degradation.\\
The performance decline observed in deep GNNs is primarily attributed to oversmoothing \cite{Li_et, JK_Nets, Klicpera, Wu}, induced by graph convolutions. Graph convolution is a type of Laplacian operator and causes node representations to converge to a stationary point when applied iteratively \cite{Li_et}. At that point, all initial information (i.e., information captured by node features) is dispersed due to the Laplacian smoothing. Therefore, oversmoothing hurts the performance of deep GNNs by making node embeddings indistinguishable across different classes.\\
Several attempts have been made to reduce oversmoothing, belonging mainly into two categories: altering the architecture of the model and modifying the topology of the graph. Approaches introducing residual and skip connections \cite{GCNII, JK_Nets} fall into the former category, while methods focusing on removing edges or nodes from the graph \cite{Dropedge, Dropnode} belong to the latter. More recent strategies explore the impact of the activation functions \cite{tanh_over, Kelesis} or leverage the Dirichlet energy \cite{dirichlet} to reduce oversmoothing.\\
Most of the existing GNN approaches train the entire model to generate node embeddings. \citet{Kipf_gcn} demonstrated that even an untrained GCN can generate informative node embeddings suitable for node classification tasks, as exemplified by their experimentation with the Zachary’s karate club network \cite{zachary}. Based on this observation, we investigate scenarios where only a single layer of the GCN is trained, while the remaining model parameters retain their initial random values. We refer to such models as partially trained ones, and we examine how altering the width of the model impacts its effectiveness in generating node embeddings. Our findings indicate that increasing the width of partially trained GCNs enhances their performance, making them resistant to oversmoothing. Our analysis provides both experimental results and theoretical insights, in order to construct a comprehensive picture of the behavior of partially trained GCNs.\\
In summary, the main contributions of this work are as follows:\\

\noindent \textbullet \textbf{Partially trained GCNs}: We investigate scenarios where only a single layer of a GCN model is permitted to receive updates during training. Our analysis reveals that as the width increases, the model is capable of obtaining high accuracy.\\

\noindent \textbullet \textbf{The power of deep partially trained GCNs:} Deep partially trained GCNs are shown experimentally to resist oversmoothing. Additionally, we highlight the advantages of these deep GNNs in scenarios with limited information, such as the ``cold start'' situation, where node features are only available for labeled nodes in a node classification task.\\

\noindent \textbullet \textbf{Position and type of the trainable layer}: We conducted experiments to explore the effect of placing the trainable layer at different positions within the network. Additionally, we examined different options for the trainable layer and found that utilizing a simple GCN layer is a good choice.\\

\noindent The paper is structured as follows. Section 2 introduces the notation and preliminaries essential for our analysis. In Section 3, we delve into our theoretical analysis. The experimental evaluation of the proposed method is detailed in Section 4, while Section 5 discusses related work. Finally, Section 6 offers a summary of our conclusions and outlines potential future directions for research. 

\section{Notations and Preliminaries}
\subsection{Notations}
We consider the task of semi-supervised node classification on a graph $G$($V,E,X$), where $|V| = N$ denotes the number of nodes, $u_i \in V$ represents the nodes, $(u_i, u_j) \in E$ stands for the edges, and $X=[x_1,...,x_N]^T \in R^{N \times C}$ indicates the initial node features, with each node having a feature vector of dimensionality $C$. The edges collectively form an adjacency matrix $A \in R^{N \times N}$, where edge $(u_i, u_j)$ is associated with element $A_{i,j}$. $A_{i,j}$ can take arbitrary real values indicating the weight (strength) of edge $(u_i, u_j)$. Node degrees are captured by a diagonal matrix $D \in R^{N \times N}$, where each element $d_i$ represents the sum of edge weights connected to node $i$. During the training phase, only the labels of a subset $V_l \subset V$ are accessible. The objective is to develop a node classifier that leverages the graph topology and the provided feature vectors to predict the label of each node.\\
\textbf{GCN}, originally proposed by \citet{Kipf_gcn}, utilizes a feed forward propagation as:
\begin{equation}\label{eq:gcn_scheme}
    H^{(l+1)} = \sigma (\hat{A}H^{(l)}W^{(l)}),
\end{equation}
where $H^{(l)} = [h^{(l)}_1,...,h^{(l)}_N]$ are node representations (or hidden vectors or embeddings) at  the $l$-th layer, with $h^{(l)}_i$ denoting the hidden representation of node $i$; $\hat{A} = \hat{D}^{-1/2} (A+I)\hat{D}^{-1/2}$ is the augmented symmetrically normalized adjacency matrix after self-loop addition, where $\hat{D}$ corresponds to the degree matrix; $\sigma(\cdot)$ is a nonlinear element-wise function, i.e. the activation function, where ReLU is a common choice in the literature, and $W^{(l)}$ is the trainable weight matrix of the $l$-th layer.\\

\subsection{Understanding Oversmoothing}
Oversmoothing occurs as node representations converge to a fixed point with increasing network depth \cite{Li_et}. At that point, node representations contain information about the graph topology, disregarding the input features. This phenomenon was initially observed in \cite{Kipf_gcn} and latter analyzed by \citet{Li_et}. The latter demonstrated that each node's new representation is a weighted average of its own representation and that of its neighbors. This mechanism is a form of Laplacian smoothing operation, which allows the node representations within each (graph) cluster, i.e. highly connected subgraph, to become more similar. Increasing similarity of neighboring node embeddings often enhances performance in semi-supervised tasks. At the same time, however, stacking multiple convolutional layers intensifies the smoothing operation, leading to oversmoothing of node representations. Consequently, the hidden representations of all nodes become similar, resulting in information loss.\\
\citet{Suzuki} have generalized the idea in \citet{Li_et} by considering that the ReLU activation function maps to a positive cone. In their analysis, oversmoothing is explained as the convergence to a subspace rather than to a fixed point. Following a similar approach, \citet{A_note} presented a different perspective to the oversmoothing problem using Dirichlet energy.\\
It is worth looking closer at the main result in \citet{Suzuki}, which estimates the speed of convergence towards a subspace $M$, where the distance between node representations tends to zero. We denote as $d_M(X)$ the distance between the feature vector $X$ and the subspace where oversmoothing is prevalent. For this distance, \citet{Suzuki} prove an interesting property.
\begin{theorem}[\citet{Suzuki}]\label{theo:1}
    Let  the largest singular value of the weight matrix $W_{lh}$ be $s_{lh}$ and $s_l = \prod \limits_{h=1}^{H_l}{s_{lh}}$, where $W_{lh}$ is the weight matrix of layer $\mathit{h}$ and $H_l$ is the network's depth, following the notation of the original paper. Then it holds that $d_M(f_l(X)) \leq s_l \lambda d_M(X)$ for any $X \in R^{N \times C}$, where f($\cdot$) is the forward pass of a GNN layer (i.e. $\sigma(AXW_{lh})$).
\end{theorem}
\noindent Theorem \ref{theo:1} indicates that increasing the network's depth results in node representations being closer to the subspace $M$. If the products of maximum singular values and the smallest eigenvalue of the Laplacian of each layer are small, then node representations asymptotically approach $M$, regardless of the initial values of node features. Extending the aforementioned theorem, the authors conclude to the following estimate about the speed of convergence to the oversmoothing subspace.
\begin{corollary}[\citet{Suzuki}]\label{col:1}
    Let $s=\sup\limits_{l \in N_+}s_{l\cdot}$ then $d_M(X^{(l)}) = O((s \lambda)^l)$, where l is the layer number and if $s\lambda <1$ the distance from oversmoothing subspace exponentially approaches zero. Where $\lambda$ is the smallest non-zero eigenvalue of I - $\hat{A}$.
\end{corollary}
\noindent According to the authors, any sufficiently deep GCN will inevitably suffer from oversmoothing, under some conditions (details can be found in \cite{Suzuki}). We build upon this result, aiming to show that partially trained GCNs resist oversmoothing and enable deeper architectures.

\subsection{Initialization and Largest Singular Value}
In our analysis, a significant portion of the model remains untrained, retaining its initial weight elements. Therefore, we briefly discuss the common weight initialization method for GNNs, i.e. Glorot initialization \cite{Xavier}. Glorot initialization proposes drawing each element of the weight matrix independently from the same zero-mean distribution, which can be either Uniform or Gaussian. For simplicity in notation and analysis, we adopt Glorot initialization using a zero-mean Gaussian distribution with variance equal to $1/d$, where $d$ represents the width of the layer.
\begin{theorem}[(Bai-Yin’s law \cite{Bai}]\label{theo:random_gauss_sing}
Let Z = $Z_{N,n}$ be an $N \times n$ random matrix with elements that are independent and identically distributed random variables having zero mean, unit variance, and finite fourth moment. Suppose that the dimensions $N$ and $n$ tend to infinity while the aspect ratio $n/N$ converges to a constant in [0, 1]. Then, almost surely, we have:
\begin{equation}\label{eq:bai-yin}
    s_{max}(Z) = \sqrt{N} + \sqrt{n} + o(\sqrt{n}).
\end{equation}
Here, $s_{max}(Z)$ denotes the largest singular value of matrix Z, and o($\cdot$) is the little--o.
\end{theorem}
\noindent Applying Bai-Yin’s law to a Glorot-initialized square matrix leads to the following Corollary.
\begin{corollary}\label{col:bai_yin}
    Let Z = $Z_{N,N}$ be a Glorot-initialized random matrix. As $N$ tends to infinity, $s_{max}(Z) \rightarrow 2 + o(1)$.\\
    This is true because Z is initialized with a zero-mean Gaussian distribution having variance equal to 1 / N, instead of unit variance (as Theorem \ref{theo:random_gauss_sing} suggests). Therefore, the largest singular value in Equation \ref{eq:bai-yin} is scaled by a factor of 1 / $\sqrt{N}$.
\end{corollary}
\noindent Corollary \ref{col:bai_yin} shows that a randomly Glorot-initialized weight matrix has a largest singular value greater than 2 as its width increases. We leverage this observation to establish a connection between our analysis and the findings in the existing literature about oversmoothing.

\subsection{Partially Trained Neural Networks}
While fully untrained models are expected to perform poorly, limited attention has been paid to partially trained networks. This approach resembles the fine-tuning \cite{pretrain} of pretrained models, which is common in large neural networks. In classical pretraining methods, a large model is initially trained on generic data and then fine-tuned on a downstream task. During fine-tuning, only a subset of its upper layers are permitted to receive updates and modify the weight values.\\
As initially noted by \citet{Kipf_gcn}, even an untrained GCN can generate node features suitable for node classification. In our analysis, we permit only a single layer of the model to receive updates during training, while maintaining the remaining layers in their initial random state.

\section{Theoretical Analysis}
\subsection{Partially Trained GCNs}
\begin{figure}[h]
    \centering
    \includegraphics[width=\columnwidth]{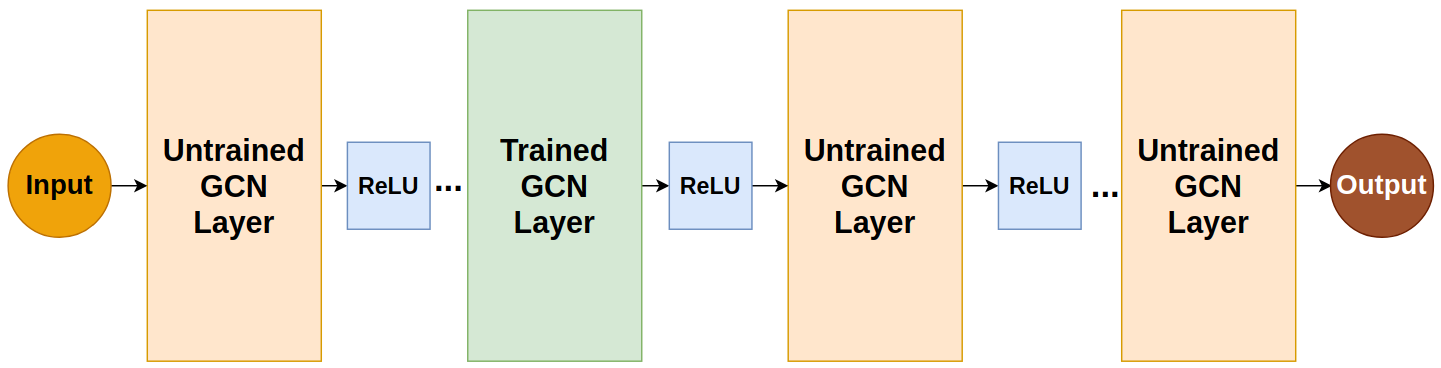}
    \caption{Architectural diagram of a partially trained GCN.}
    \label{fig:architecture}
\end{figure}

We propose partially training of a GCN, where the training process is constrained to a single layer of the network. The model architecture remains the same as the one presented in Equation \ref{eq:gcn_scheme}, but all weight matrices except one remain constant during training. The trainable layer can be placed anywhere within the network. The proposed architecture is presented in Figure \ref{fig:architecture}. With training limited to a single layer (i.e., $j$-th layer) of the network, we can partition the weight matrices into three sets: i) The lower $j-1$ untrained matrices, ii) The $j$-th trainable matrix, and iii) The upper $L-j$ untrained matrices, where $L$ is the model depth. Since learning is a dynamic process, which cannot be fully controlled, we investigate the properties of the two sets containing the untrained matrices and their effect on the model's behavior.

\vspace{0.6cm}
\noindent To simplify the notation, we omit the influence of the ReLU activation functions in our analysis. This simplification also provides an upper bound for the final node representations of the model, due to the fact that ReLU is 1-Lipschitz. That particular property restricts the extent to which the output of the function can change in response to changes in its input. Therefore, by disregarding the impact of ReLU activation functions, we establish a theoretical ceiling on the improvement achievable in the model's node representations.\\
Considering the partition of matrices into three sets and disregarding the ReLUs, we rewrite Equation \ref{eq:gcn_scheme} to represent the final node embeddings, as follows:
\begin{equation*}
    H^{(L)}=\hat{A}^{L-1} \cdot X\bigg(\cdot \prod\limits_{i=1}^{j-1}{W^{(i)}}\bigg) \cdot W^{(j)} \cdot \bigg(\prod\limits_{i=j+1}^{L-1}{W^{(i)}}\bigg) = 
\end{equation*}
\begin{equation}\label{eq:partially_untrained}
    \hat{A}^{L-1} \cdot X \cdot W^{left} \cdot W^{(j)} \cdot W^{right},
\end{equation}
where $L$ is model's depth, $\hat{A}$ is the augmented symmetrically normalized adjacency matrix, $X$ are the initial node features, $W^{(j)}$ is the weight matrix of the trainable layer of the model, and $W^{left}, W^{right}$ denote the products of the untrained weight matrices.\\
Depending on the value of $j$, representing the position of the trainable layer, the behavior of the network changes: i) $j = 1$: Learning occurs in the first layer, with subsequent layers propagating and aggregating information using random weights. ii) $j \in [2,L-2]$: Initial node features undergo random transformations, followed by learning in layer $j$. The learned features are then randomly aggregated by subsequent layers. iii) $j=L-1$: Initial features pass through the random network, where information is propagated and aggregated. The final layer is responsible for generating meaningful representations.\\
In the following analysis, we study the formulation of $W^{left} \text{ and } W^{right}$, the distribution of their elements, and how they affect the information flow within the network.

\subsection{Product of Untrained Weight Matrices}
We now take a closer look at the properties of $W^{left}$ and $W^{right}$. Our analysis focuses on the product of two untrained matrices, denoted as $W^{(1)}$ and $W^{(2)}$, of size $d \times d$. The product of these two matrices is expressed as:

\begin{equation*}
    W^{(1)} \cdot W^{(2)} = \left[\begin{array}{ccc}
       w^1_{1,1}  & \hdots & w^1_{1,d}\\
       \vdots  & \ddots & \vdots\\
       w^1_{d,1}  & \hdots & w^1_{d,d}\\
    \end{array}\right] \cdot \left[\begin{array}{ccc}
       w^2_{1,1}  & \hdots & w^2_{1,d}\\
       \vdots  & \ddots & \vdots\\
       w^2_{d,1}  & \hdots & w^2_{d,d}\\
    \end{array}\right] =\\
\end{equation*}
\begin{equation*}
    =\left[\begin{array}{ccc}
       w^{1,2}_{1,1}  & \hdots & w^{1,2}_{1,d}\\
       \vdots  & \ddots & \vdots\\
       w^{1,2}_{d,1}  & \hdots & w^{1,2}_{d,d}\\
    \end{array}\right].
\end{equation*}
We focus on $w^{1,2}_{1,1}$ of the resulting matrix, as the same analysis extends to all elements. Since both $W^{(1)}$ and $W^{(2)}$ remain constant during training, and all their elements are drawn from the same distribution, it suffices to determine the distribution of $w^{1,2}_{1,1}$. We are interested in the distribution of $w^{1,2}_{\cdot,\cdot}$, because they remain fixed throughout the training process, thereby influencing the flow of the initial node information within the model. By characterizing their distribution, we aim to gain insights into how this information evolves and study the properties of the resulting embeddings.\\
The value of $w^{1,2}_{1,1}$ is computed as:
\begin{equation}\label{eq:element_format}
    w^{1,2}_{1,1} = \sum\limits_{i=1}^d{w^1_{1,i} w^2_{i,1}}.
\end{equation}
Each term of the sum in Equation \ref{eq:element_format} is a product of two independent and identically distributed Gaussian Random Variables (R.V.). The following Lemma shows the distribution of the product of two such R.V.s. 
\begin{lemma}\label{lemma:rv_prod}
    Let $Y_1, Y_2$ be two independent R.V.s which follow the same Gaussian distribution. Their product can be written as:
    \begin{equation*}
        Y_1Y_2 = \frac{(Y_1+Y_2)^2}{4} - \frac{(Y_1-Y_2)^2}{4}.
    \end{equation*}
    If $(Y_1 \pm Y_2)$ follow a Normal distribution then $(Y_1 \pm Y_2)^2$ will follow a Chi-square $\left(\chi^2\right)$ distribution. Therefore,
    \begin{equation*}
        Y_1Y_2 \sim \frac{Var(Y_1+Y_2)}{4}Q - \frac{Var(Y_1-Y_2)}{4}R =
    \end{equation*}
    \begin{equation*}
        \frac{Var(Y_1) + Var(Y_2)}{4}(Q-R),
    \end{equation*}
    where $Q, R \sim \chi^2_1$ and are independent of each other.
\end{lemma}
\noindent Lemma \ref{lemma:rv_prod} shows that each term in the sum of Equation \ref{eq:element_format} follows a $\chi^2_1$ distribution with one degree of freedom. Based on that result, we focus on determining the distribution of the difference of two R.V.s, when each of them follows a $\chi^2_1$ distribution.

\begin{lemma}\label{lemma:variance_gamma}
    Let $F = Q - R$, where $Q,R \sim \chi^2_1$. Then $F$ follows a symmetric around zero Variance-Gamma ($V.G._{(\alpha, \beta, \lambda, \mu)}$) distribution with parameters: $\alpha = \lambda = 1/2$ and $\beta = \mu = 0$.
\end{lemma}
\noindent The proof is based on the moment-generating function of the Chi-square distribution with one degree of freedom.
\begin{equation*}
    M_Q(t) = M_R(t) = (1 - 2t)^{-1/2}.
\end{equation*}
\begin{equation}\label{eq:moment_gamma}
    M_F(t) = \left(\frac{1/4}{1/4-t^2}\right)^{1/2}.
\end{equation}
Equation \ref{eq:moment_gamma} shows that the moment of F matches the moment of a variance-gamma distribution with the aforementioned parameters.\\
\noindent Using the parameter set defined in Lemma \ref{lemma:variance_gamma} to the equations of the Variance-Gamma distribution we get: i) $E[F] = 0$, and ii) $Var(F) = 4$.\\
Combining the results of Lemma \ref{lemma:rv_prod} and Lemma \ref{lemma:variance_gamma}, we deduce that the product of two independent R.V.s, $Y_1,$ and $Y_2$, each following a Gaussian distribution, follows a Variance-Gamma distribution. Specifically, the parameter set of the distribution is outlined in Lemma \ref{lemma:variance_gamma}, i.e. $Y_1Y_2 \sim J_{Y_1,Y_2} \cdot F$, where $J_{Y_1,Y_2} = \big(Var(Y_1) + Var(Y_2)\big) / 4$, and $F \sim V.G._{(1/2, 0, 1/2, 0)}$.\\
Applying this result to Equation \ref{eq:element_format} leads to:
\begin{equation*}
    w^{1,2}_{1,1} \sim \frac{Var(Init(W^{(1)})) + Var(Init(W^{(2)}))}{4} \sum\limits_{i=1}^d{F_i},
\end{equation*}
where $Var(Init(W^{(k)}))$ is the variance of the distribution used for the initialization of the weight matrix of the $k$-th layer, and  $F_i \sim V.G._{(1/2, 0, 1/2, 0)}$.\\
Both $W^{(1)}$ and $W^{(1)}$ are initialized by a zero-mean Gaussian distribution with variance equal to $1/d$ (Glorot initialization), which in turn leads to the following result about the distribution of each element of the $W^{(1)} \cdot W^{(2)}$ matrix:
\begin{equation*}
w^{1,2}_{1,1} \sim \frac{1/d + 1/d}{4} \sum\limits_{i=1}^d{F_i} = \frac{1}{2d} \sum\limits_{i=1}^d{F_i} = \frac{1}{2}\Bar{F_d},    
\end{equation*}
with $\Bar{F_d}$ being the average of $F_i$'s.\\
For the last step of our analysis of the product of the weight matrices, we utilize the Central Limit Theorem (CLT). CLT states that the distribution of $\sqrt{d}(\Bar{F_d} - E[F_i]) \xrightarrow[]{approx.} \mathcal{N}(0,Var(F_i))$ as $d$ increases.\\
Considering that $E[F_i] = 0$ and $Var(F_i)=4$, we conclude that the distribution of $\Bar{F_d} \xrightarrow[]{approx.} \mathcal{N}\left(0,\frac{4}{d}\right)$.\\
Hence, we arrive at the distribution of  $w^{1,2}_{1,1} \sim \frac{1}{2}\Bar{F_d} \xrightarrow[]{approx.} \mathcal{N}\left(0, \frac{1}{d}\right)$, which is the same distribution as the one used in Glorot initialization. Consequently, the elements of the product of two weight matrices follow the same distribution as that followed by the elements of the initial matrices. Using induction we can also prove that the product of an arbitrary number of untrained weight matrices is a Gaussian matrix, with elements following a $\mathcal{N}\left(0, \frac{1}{d}\right)$ distribution, as model's width (i.e., $d$) increases.

\subsection{Effect of $W^{left}$}\label{sec:effect_untrain}
We have proved that both $W^{left}$ and $W^{right}$ are Gaussian matrices, with random elements following the distribution of the Glorot initialization method. Looking back at Equation \ref{eq:partially_untrained}, since both the graph topology and the initial node features are fixed, their product ($B=\hat{A}^{L-1}X$) remains constant during training. Therefore, we investigate the effect of multiplying a Gaussian matrix with a constant matrix, i.e. matrix $B$. In order to keep the notation simple, we analyze the case of $j=L-1$ in Equation \ref{eq:partially_untrained}, which means that $W^{right}$ disappears. Equation \ref{eq:partially_untrained} is transformed as follows:\\
\begin{equation}\label{eq:effect_of_w_left}
    H^{(L)} = \big(B \cdot W^{left} \big) \cdot W^{(j)}.
\end{equation}

\begin{lemma}\label{lemma:prod_distrib_matrix}
    Let $Y$ be a random vector with $Y \sim \mathcal{N}(\mu, \Sigma)$, and $O$ a non-singular matrix, the distribution of $O \cdot Y$ is given as:
    \begin{equation*}
        O \cdot Y \sim \mathcal{N}(O\mu, O\Sigma O^T).
    \end{equation*}
\end{lemma}
\noindent According to Lemma \ref{lemma:prod_distrib_matrix}, every column of the product of a Gaussian matrix and a non-singular matrix follows the same Gaussian distribution. Hence the resulting matrix also follows that particular Gaussian distribution.\\
The proof of Lemma \ref{lemma:prod_distrib_matrix} comes from the definition of a multivariate Gaussian random vector and the impact of an affine transformation on its distribution. Lemma \ref{lemma:prod_distrib_matrix} indicates that $B \cdot W^{left} \sim \mathcal{N}(0, \frac{1}{d}BB^T)$, because $W^{left} \sim \mathcal{N}(0, \frac{1}{d})$ and $B$ is non-singular.\\
Let us take a closer look at matrix $B$, ignoring temporarily $W^{left}$. Matrix $B$ comprises node features that are averaged multiple times, based on the underlying graph topology. As the depth increases, the number of copies of $\hat{A}$ in $B$ also increases, intensifying the averaging process. This, in turn, increases the similarity and the correlation between the feature vectors (i.e., rows of $B$). Highly correlated features become too similar, causing oversmoothing.\\
However, the presence of $W^{left}$ in Equation \ref{eq:effect_of_w_left} reduces that correlation. The product $B \cdot W^{left}$ follows a Gaussian distribution with a predefined variance, i.e., $\frac{1}{d} B B^T = \frac{N}{d} (\frac{1}{N} B B^T)$. Without loss of generality, we assume that the initial node features (i.e., rows of $X$) are drawn from a zero-mean distribution, whose covariance matrix is unknown. Hence, matrix $B$ has also rows drawn from a zero-mean distribution with unknown covariance matrix. The term $\frac{1}{N}B B^T$ is an estimator of the covariance matrix of that particular distribution, followed by the rows of $B$, which are highly correlated, as stated above.\\
Since the rows of $B$ are highly correlated, as the depth increases, we expect that the covariance matrix (and its estimator) will contain large elements. This, in turn, highlights the role of the term $\frac{N}{d}$, which tones down these large elements. Considering the above statement, we would like to adjust the elements of the covariance matrix, in order to: i) prevent similar embeddings for all nodes, and ii) avoid highly dissimilar embeddings between nodes of the same class. Therefore, the term $\frac{N}{d}$ plays a crucial role in reducing the correlation between the rows of matrix $B$. Given that $N$ is fixed, we can reduce the correlation, by increasing $d$.\\
This observation highlights the effect of $W^{left}$, which can increase the dissimilarity between node embeddings, counteracting the oversmoothing caused by repeated multiplications with the adjacency matrix. Finally, we observe that $d$ serves a dual purpose, as its increase (a) strengthens the conditions of CLT; namely the more random variables we sum, the stronger is the convergence of their average to a Gaussian distribution, and (b) reduces the correlation between node features (as explained above). However, excessively increasing $d$ might lead to very dissimilar node embeddings or poor training performance. 

\subsection{Untrained Weight Matrices and Oversmoothing}
Conceptually, as discussed in subsection \ref{sec:effect_untrain}, untrained weight matrices act as reducers of the correlation between node embeddings before learning occurs. Their presence helps the model to preserve the informative aspects of the initial node embeddings, while also aggregating information from distant nodes in the multi-hop neighborhood, due to the powers of the adjacency matrix. Consequently, the trainable layer is applied to node embeddings that contain information from distant nodes, thereby enabling effective learning.\\
We now shift our focus to the singular values of the weight matrices, which are known to influence oversmoothing. The untrained matrices are all initialized using Glorot method and remain constant during training, which means that their singular values do not change. Leveraging Corollary \ref{col:bai_yin} about these matrices, we conclude that their largest singular value is greater than 2, as their width (i.e., $d$) tends to infinity. Therefore, the product of the largest singular values of the untrained weight matrices maintains a high value. According to Corollary \ref{col:1}, a large value of the product of the largest singular values of the weight matrices allows the model to reduce oversmoothing, which is what we expect from the partially trained GCNs.

\section{Experiments}

\begin{figure*}
    \centering
    \includegraphics[scale=.46]{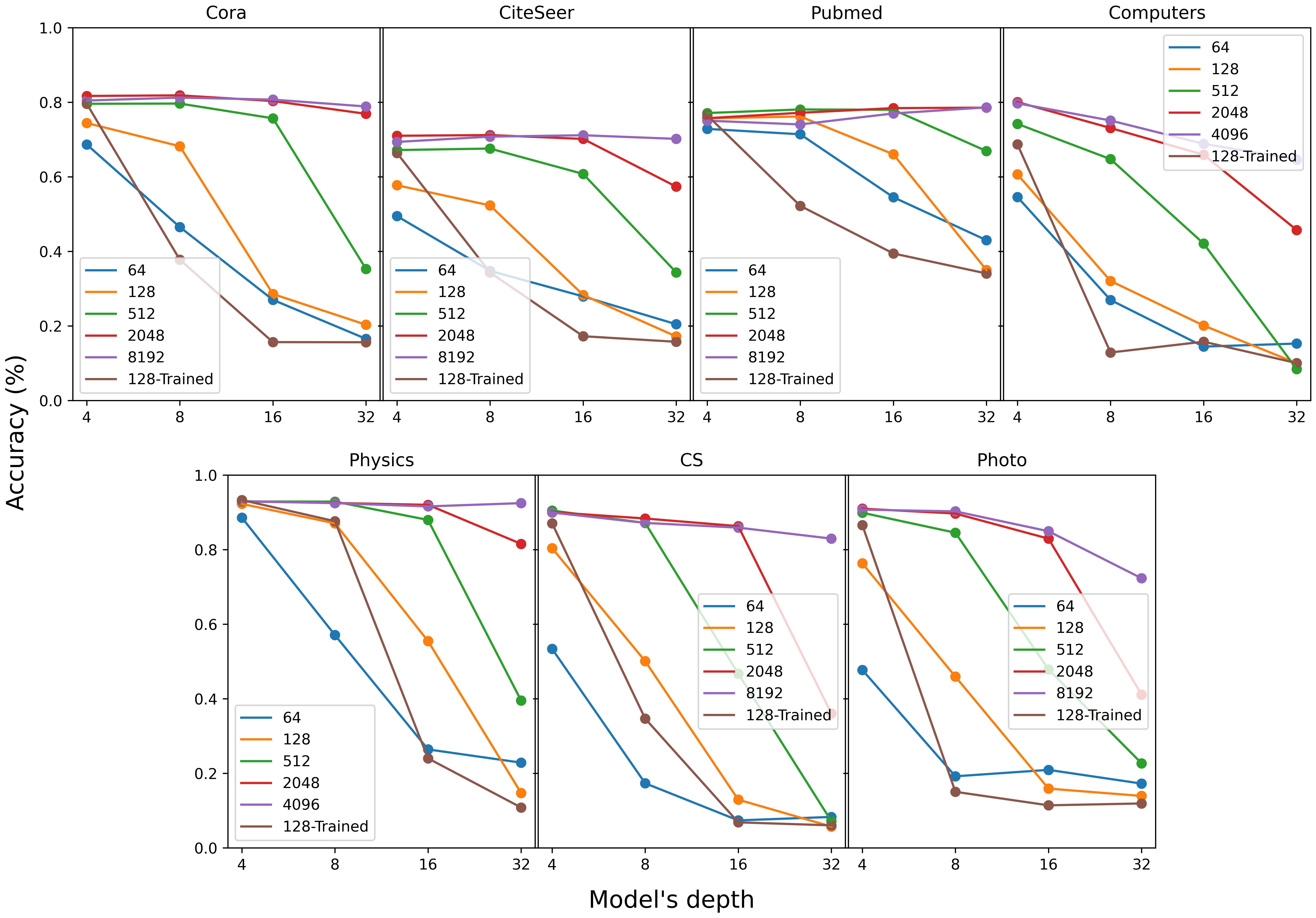}
    \caption{Comparison between a fully trained GCN and 5 different configurations (in terms of width) of partially trained GCNs across 7 datasets for varying depth. The trainable layer is always the second.}
    \label{fig:vary_width_depth}
\end{figure*}

\subsection{Experimental Setup}\label{sec:exp_setup}
\textbf{Datasets:} We conduct experiments on well-known benchmark datasets for node classification: \textit{Cora, CiteSeer, Pubmed}, utilizing the same data splits as in \cite{Kipf_gcn}, where all nodes except the ones used for training and validation are used for testing. Furthermore, we use the \textit{Photo, Computer, Physics} and \textit{CS} datasets, adopting the splitting method presented in \cite{rest_datasets}. \\
\textbf{Models:} The focus of our work is on the effect of the model's width, under the partially trained setting. Therefore, we use the conventional GCN architecture \cite{Kipf_gcn} to generate models. These models update a single layer, and keep the rest of the model weights frozen to the initial random values.\\
\textbf{Hyperparameters:} We vary the width of GCNs between 64 and 8192. The fully trained models are subject to $L_2$ regularization with a penalty of $5 \cdot 10^{-4}$, while the partially trained GCNs do not use regularization. The learning rate for the fully trained models is set to $10^{-3}$, whereas for the partially trained, it is set to 0.1. The above values are determined, based on the performance of the models on the validation set.\\
\textbf{Configuration:} Each experiment is run 10 times and we report the average accuracy and standard deviation over these runs. We train all models for 200 epochs using Cross Entropy as the loss function.

\subsection{Experimental Results} 
\textbf{One trainable layer for different widths and depths:}\\
In our first experiment, we investigate the effect of width and depth in partially trained GCNs. Specifically, in Figure \ref{fig:vary_width_depth} we observe that wider networks tend to be more resistant to oversmoothing, as the depth of the network increases. This aligns with our theoretical findings, as presented in section 3. The trainable layer of the networks is the second, a choice which based on experimentation. Additionally, partially trained GCNs outperform fully trained ones given a wide enough trainable layer. Therefore, a wide single layer can effectively learn informative node representations, when node information is appropriately aggregated by untrained layers. It is noteworthy, that further increasing the width of fully trained GCNs does not improve their performance, while, in contrast, we observe significant improvements for the partially trained models as their width increases.

\begin{table}
\centering
\caption{Comparison of different model widths with trained GCN, in the “cold start” scenario. Only the features of the nodes in the training set are available to the model. We present the best accuracy of the model and the depth (i.e. \# Layers) this accuracy is achieved.}
\vspace{0.6cm}
\begin{tabular}{|c|c|c c|}
        \hline
        Dataset & Width & Accuracy (\%) \& std & \#L\\
        \hline
        \multirow{4}{*}{Cora} &
        128 & 63.31 \scriptsize$\pm$ 2.5 & 6 \\
        {} & 512 & 70.60 \scriptsize$\pm$ 1.5 & 15 \\
        {} & 8192 & \textbf{73.24 \scriptsize$\pm$ 0.6} & 17 \\
        {} & Trained & 64.86 \scriptsize$\pm$0.7 & 4 \\
        \hline
        \multirow{4}{*}{CiteSeer} &
        128 & 42.09 \scriptsize$\pm$ 0.7 & 4 \\
        {} & 512 & 48.17 \scriptsize$\pm$ 1.2 & 14 \\
        {} & 8192 & \textbf{51.32 \scriptsize$\pm$ 0.6} & 31 \\
        {} & Trained & 41.95 \scriptsize$\pm$0.2 & 4 \\
        \hline
        \multirow{4}{*}{Pubmed} &
        128 & 64.68 \scriptsize$\pm$ 3.2 & 7 \\
        {} & 512 & 69.80 \scriptsize$\pm$ 0.8 & 7 \\
        {} & 8192 & \textbf{72.73 \scriptsize$\pm$ 0.5} & 16 \\
        {} & Trained & 64.54 \scriptsize$\pm$0.9 & 4 \\
        \hline
        \multirow{4}{*}{Physics} &
        128 & 38.61 \scriptsize$\pm$ 0.0 & 2 \\
        {} & 512 & 75.82 \scriptsize$\pm$ 0.6 & 4 \\
        {} & 4096 & 88.11 \scriptsize$\pm$ 0.5 & 14 \\
        {} & Trained & \textbf{94.00 \scriptsize$\pm$0.1} & 2 \\
        \hline
        \multirow{4}{*}{CS} &
        128 & 40.52 \scriptsize$\pm$ 0.2 & 2 \\
        {} & 512 & 63.94 \scriptsize$\pm$ 0.8 & 4 \\
        {} & 8192 & 77.49 \scriptsize$\pm$ 0.4 & 14 \\
        {} & Trained & \textbf{89.95 \scriptsize$\pm$0.2} & 1 \\
        \hline
        \multirow{4}{*}{Photo} &
        128 & 73.87 \scriptsize$\pm$ 0.5 & 2 \\
        {} & 512 & 76.72 \scriptsize$\pm$ 0.0 & 2 \\
        {} & 8192 & \textbf{86.11 \scriptsize$\pm$ 0.2} & 5 \\
        {} & Trained & 83.19 \scriptsize$\pm$5.1 & 5 \\
        \hline
        \multirow{4}{*}{Computers} &
        128 & 61.07 \scriptsize$\pm$ 0.7 & 2 \\
        {} & 512 & 67.07 \scriptsize$\pm$ 0.0 & 2 \\
        {} & 4096 & \textbf{76.58 \scriptsize$\pm$ 0.4} & 5 \\
        {} & Trained & 68.83 \scriptsize$\pm$6.5 & 4 \\
        \hline
    \end{tabular}
    \label{tab:cold_start}
\end{table}

\begin{table*}[ht]
\centering
\caption{Performance comparison of GCN models with different width and depth, as well as different placement of the trainable layer.}
\vspace{0.6cm}
\begin{tabular}{|c|c|c|c|c|c|c|}
        \hline
        \multicolumn{7}{|c|}{Accuracy (\%) \& std}\\
        \hline
        \multirow{2}{*}{Dataset} & \multirow{2}{*}{(Width, Depth)} & 
        \multicolumn{5}{|c|}{Position} \\
        \cline{3-7}
        {} & {} & 2 & 4 & 8 & 16 & 32\\
        \hline
        \multirow{6}{*}{CiteSeer} & 
        (512, 8) & \textbf{67.58 \scriptsize$\pm$ 1.25} & 62.13 \scriptsize$\pm$ 1.18 & 53.47 \scriptsize$\pm$ 1.85 & - & -\\
        \cline{2-7}
        {} & (2048, 8) & \textbf{71.20 \scriptsize$\pm$ 0.49} & 67.20 \scriptsize$\pm$ 1.18 & 61.14 \scriptsize$\pm$ 1.11 & - & -\\
        \cline{2-7}
        {} & (8192, 8) & \textbf{70.78 \scriptsize$\pm$ 1.45} & 69.52 \scriptsize$\pm$ 1.08 & 66.39 \scriptsize$\pm$ 0.74 & - & -\\
        \cline{2-7}
        {} & (512, 32) & 34.35 \scriptsize$\pm$ 12.25 & 32.40 \scriptsize$\pm$ 9.81 & 29.01 \scriptsize$\pm$ 9.86 & 27.72 \scriptsize$\pm$ 11.78 & \textbf{30.12 \scriptsize$\pm$ 10.94}\\
        \cline{2-7}
        {} & (2048, 32) & 57.36 \scriptsize$\pm$ 6.46 & 55.76 \scriptsize$\pm$ 5.72 & 57.22 \scriptsize$\pm$ 2.87 & 56.72 \scriptsize$\pm$ 3.84 & \textbf{58.69 \scriptsize$\pm$ 3.80}\\
        \cline{2-7}
        {} & (8192, 32) & \textbf{70.18 \scriptsize$\pm$ 0.64} & 68.21 \scriptsize$\pm$ 0.87 & 65.45 \scriptsize$\pm$ 1.06 & 65.45 \scriptsize$\pm$ 0.99 & 65.99 \scriptsize$\pm$ 1.09\\
        \hline
        \multirow{6}{*}{Photo} & 
        (512, 8) & \textbf{84.53 \scriptsize$\pm$ 1.32} & 83.54 \scriptsize$\pm$ 0.74 & 66.85 \scriptsize$\pm$ 5.44 & - & -\\
        \cline{2-7}
        {} & (2048, 8) & \textbf{89.68 \scriptsize$\pm$ 0.12} & 89.41 \scriptsize$\pm$ 0.18 & 82.59 \scriptsize$\pm$ 0.36 & - & -\\
        \cline{2-7}
        {} & (8192, 8) & \textbf{90.26 \scriptsize$\pm$ 0.21} & 89.82 \scriptsize$\pm$ 0.22 & 87.73 \scriptsize$\pm$ 0.18 & - & -\\
        \cline{2-7}
        {} & (512, 32) & 22.69 \scriptsize$\pm$ 15.00 & 22.15 \scriptsize$\pm$ 11.70 & 24.42 \scriptsize$\pm$ 15.12 & \textbf{25.68 \scriptsize$\pm$ 15.29} & 21.61 \scriptsize$\pm$ 12.61\\
        \cline{2-7}
        {} & (2048, 32) & 41.12 \scriptsize$\pm$ 17.04 & 44.96 \scriptsize$\pm$ 15.92 & 50.78 \scriptsize$\pm$ 13.71 & \textbf{51.76 \scriptsize$\pm$ 13.70} & 50.19 \scriptsize$\pm$ 13.14\\
        \cline{2-7}
        {} & (8192, 32) & \textbf{72.32 \scriptsize$\pm$ 2.93} & 71.39 \scriptsize$\pm$ 2.04 & 70.58 \scriptsize$\pm$ 1.27 & 70.66 \scriptsize$\pm$ 1.95 & 71.62 \scriptsize$\pm$ 1.80\\
        \hline
        \multirow{6}{*}{Computers} & 
        (512, 8) & \textbf{64.74 \scriptsize$\pm$ 2.14} & 63.81 \scriptsize$\pm$ 1.57 & 59.70 \scriptsize$\pm$ 1.03 & - & -\\
        \cline{2-7}
        {} & (2048, 8) & \textbf{73.10 \scriptsize$\pm$ 0.26} & 72.22 \scriptsize$\pm$ 0.22 & 64.77 \scriptsize$\pm$ 0.71 & - & -\\
        \cline{2-7}
        {} & (4096, 8) & \textbf{74.80 \scriptsize$\pm$ 0.00} & 73.15 \scriptsize$\pm$ 0.00 & 66.52 \scriptsize$\pm$ 0.00 & - & -\\
        \cline{2-7}
        {} & (512, 32) & 8.43 \scriptsize$\pm$ 5.18 & 8.83 \scriptsize$\pm$ 5.24 & \textbf{12.71 \scriptsize$\pm$ 11.89} & 10.82 \scriptsize$\pm$ 5.48 & 8.33 \scriptsize$\pm$ 5.21\\
        \cline{2-7}
        {} & (2048, 32) & 45.74 \scriptsize$\pm$ 19.44 & 48.15 \scriptsize$\pm$ 19.88 & 49.38 \scriptsize$\pm$ 19.14 & 50.85 \scriptsize$\pm$ 17.11 & \textbf{50.97 \scriptsize$\pm$ 17.32}\\
        \cline{2-7}
        {} & (4096, 32) & 65.00 \scriptsize$\pm$ 0.00 & \textbf{67.62 \scriptsize$\pm$ 0.00} & 65.17 \scriptsize$\pm$ 0.00 & 66.28 \scriptsize$\pm$ 0.00 & 64.84 \scriptsize$\pm$ 0.00\\
        \hline
    \end{tabular}
    \label{tab:pos_inv}
\end{table*}
\noindent \textbf{Partially trained GNNs can solve the ``cold start'' problem:}\\
The second set of experiments aimed at demonstrating the advantages of using deep architectures that are resistant to oversmoothing. For this purpose, we simulate the ``cold start'' scenario, which resembles the situation where a new item enters a recommender system without any features. This setup, initially proposed in \cite{Pairnorm}, requires leveraging information from distant nodes to create meaningful embeddings for the new item. In order to simulate it, we create the cold-start variants of the datasets, by removing feature vectors from the unlabeled nodes and replacing them with all-zero vectors. These modified datasets enable us to assess the effectiveness of deep architectures in handling scenarios with limited feature information.\\
For different widths of partially trained GCNs, we present the best performance achieved and the depth at which the model attains that performance in Table \ref{tab:cold_start}.\\
In this context, we observed that partially trained GCNs perform better if both of their last two layers are allowed to receive updates. This is due to the increased difficulty of the problem, compared to standard node classification without missing features. From Table \ref{tab:cold_start}, we observe that the deeper models consistently outperform shallower ones in almost every dataset, underscoring the necessity of relatively deep architectures, which are less prone to oversmoothing. Additionally, table \ref{tab:cold_start} shows that as width increases, the performance of the models improves, and the depth at which optimal performance is attainable also increases. It is worth noting that the fully trained GCNs generally exhibit inferior performance, which is also attained by shallower architectures. This is due to the high-degree of oversmoothing of such models, which makes deeper architectures unusable. There are two exceptions to this trend: the \textit{Physics} dataset, where hardware limitations prevented the use of very wide networks, and the \textit{CS} dataset, where the ``cold start'' scenario seems to be adequately addressed by a 1-layer fully-trained model.\\
\textbf{Role of the position of the trainable layer:}\\
Having demonstrated the advantages of partially trained GCNs in both standard node classification tasks and the ``cold start'' problem, we now conduct a set of ablation experiments. According to Equation \ref{eq:partially_untrained}, the trainable layer can be placed at any position within the network. Our theoretical analysis has primarily focused on the scenario where the final layer is trainable.\\
Table \ref{tab:pos_inv} shows the impact of varying the position of the trainable layer across different combinations of widths and depths. We observe that, in general, the optimal position for the trainable layer is the second. Placing the trainable layer at a relatively early stage of the model, facilitates training on the basis of a very local aggregation of information, i.e. 2-hop. Subsequent convolutional layers then distribute the learned embeddings, while also preventing over-correlation, as explained theoretically. The performance gap between different positions of the trainable layer diminishes with increasing width, showing that a large width makes the model more robust to the placement of the trainable layer. Additionally, the optimal position is different for different datasets. Table \ref{tab:pos_inv} presents the results for three datasets chosen to highlight this observation. For the rest of the datasets we have observed similar behavior and the respective results appear in the Appendix.\\
\textbf{MLP vs GCN for trainable layer:}\\
In this experiment, we aimed to answer the question of whether the trainable layer has to be a graph convolutional layer, or if it could be replaced by a Multi-Layer Perceptron (MLP). The underlying hypothesis was that if sufficient information has already been effectively propagated through the network by the untrained graph convolution layers, then an MLP could potentially generate informative node representations.\\
Table \ref{tab:mlp_gcn} presents the performance achieved by models employing either a single GCN or an MLP as their trainable layers. We focus on a rather challenging configuration of a 16-layer model, which has access to distant information but needs to avoid oversmoothing. Results indicate that GCN slightly outperforms MLP in terms of accuracy and stability, as evidenced by achieving better scores with smaller standard deviations. However, the performance improvement is marginal, suggesting that both approaches can generate informative node representations, provided the necessary information has been appropriately received and not excessively mixed by graph convolution.\\
\begin{table}
    \centering
    \caption{Performance comparison between an MLP and a GCN, used as trainable layers, in a 16-layer GCN model, for \textit{Cora, Citeseer} and \textit{Pubmed} datasets.}
    \vspace{0.6cm}
    \begin{tabular}{|c|c|c|}
        \hline
        \multicolumn{3}{|c|}{Accuracy (\%) \& std}\\
        \hline
        \diagbox{Dataset}{Model} & MLP & GCN \\
        \hline
        Cora & 80.83 \scriptsize$\pm$ 0.8 & \textbf{81.09 \scriptsize$\pm$ 0.4}\\
        \hline
        CiteSeer & 70.31 \scriptsize$\pm$ 1.2 & \textbf{71.15 \scriptsize$\pm$ 1.1}\\
        \hline
        Pubmed & 78.32 \scriptsize$\pm$ 0.8 & \textbf{78.55 \scriptsize$\pm$ 0.3} \\
        \hline
    \end{tabular}
    \label{tab:mlp_gcn}
\end{table}

\noindent \textbf{Varying width within the network:}\\
In our final experiment, we investigate the impact of changing the width of the untrained layers. Table \ref{tab:vary_width} shows the performance of a 16-layer partially trained GCN with varying width across all of its untrained layers. Each row corresponds to a model with a different configuration. In our experiments, wide layers are placed lower in the model, but changing their position does not yield improvements.\\
Results indicate that models having the majority of their layers wide, perform better. This observation is aligned with our theoretical analysis, in which we have assumed wide square weight matrices (i.e., maintaining the same width throughout the network).

\begin{table}
    \centering
    \caption{Performance comparison on the \textit{Cora} dataset, when we vary the width of a 16-layer GCN model, in which only the final layer is trainable. For each configuration each width column indicates the number of the untrained layers which have that particular width. Wider layers are placed at the lowest positions.}
    \vspace{0.6cm}
    \begin{tabular}{|c|c|c|c|}
        \hline
        \multicolumn{3}{|c|}{Width} & \multirow{2}{*}{Accuracy (\%) \& std}\\
        \cline{1-3}
        2048 & 4096 & 8192 & \\
        \hline
        12 & 0 & 3 & 76.54 \scriptsize$\pm$ 1.0\\
        \hline
        8 & 0 & 7 & 76.82 \scriptsize$\pm$ 1.0\\
        \hline
        4 & 0 & 11 & 77.50 \scriptsize$\pm$ 1.0\\
        \hline
        0 & 0 & 15 & \textbf{79.45 \scriptsize$\pm$ 1.0}\\
        \hline
        0 & 12 & 3 & 78.37 \scriptsize$\pm$ 0.7\\
        \hline
        0 & 8 & 7 & 78.62 \scriptsize$\pm$ 1.0\\
        \hline
        0 & 4 & 11 & 79.06 \scriptsize$\pm$ 0.8\\
        \hline
    \end{tabular}
    \label{tab:vary_width}
\end{table}

\section{Related Work}
There is limited published work on partially trained graph neural networks, despite the observation made in the original paper by \citet{Kipf_gcn}. They observed that an untrained GCN can indeed generate meaningful node representations, but did not explore the underlying reasons or mechanisms. They also did not study the effect of model width in such models.\\
An area of research that is marginally related to ours is the study of Neural Tangent Kernels (NTK), which formulates the relationship between the width of deep neural networks and kernel methods. NTK provides insights into the dynamics of neural network training using a system of differential equations, particularly in the limit of infinite width, where the network behavior becomes increasingly deterministic. While NTK has primarily been applied to deep neural networks, recent work has extended its applicability to GNNs \cite{NTK}. However, the focus of NTK literature is on the theoretical understanding of graph classification tasks. In contrast, we examine mainly node classification tasks and further establish a connection between the behavior of these models and oversmoothing.\\
Finally, another relevant area of research is the ongoing efforts to reduce oversmoothing in GNNs. Several approaches have been proposed to address the problem, typically involving modifications to either the model architecture or the structure of the underlying graph \cite{GCNII, Dropedge}. \citet{GCNII} proposed the use of residual connections and identity mapping in order to enable deep architectures and alleviate oversmoothing. That particular approach injects part of the initial information into higher layers of the network, implicitly downgrading the importance of intermediate layers, and forces the model to remain in the local neighborhood around each node. DropEdge \cite{Dropedge} aimed to address oversmoothing by altering the graph topology, and removing edges at random, in order to slow down message passing. In contrast to these approaches, our method maintains the original graph topology and uses the prominent GCN architecture, keeping the larger part of the model untrained and showing its capabilities in node classification tasks, even in the demanding ``cold start'' scenario.

\section{Conclusion}
In this work we proposed the use of partially trained GCNs for node classification tasks. We have shown theoretically that training only a single layer in wide GCNs is sufficient to achieve comparable or superior performance to fully trained models. Additionally, we linked our proposed method to existing literature on oversmoothing and demonstrated that partially trained GCNs can perform well even as their depth increases. We emphasized the dual role of width in our analysis, both in terms of Central Limit Theorem convergence and for reducing the correlation of node embeddings, due to repeated weight averaging. We have assessed our theoretical results through a variety of experiments that confirmed the potential of partially trained GCNs. As future work, we aim to explore the impact of untrained layers on more complex architectures, incorporating attention mechanisms, residual and skip connections. Finally, we would like to shed more light to the dataset properties and the number of trainable layers that are required in the partially trained setup.



\begin{ack}
The research work was supported by the Hellenic Foundation for Research and Innovation (HFRI) under the 4th Call for HFRI PhD Fellowships (Fellowship Number: 10860).
\end{ack}



\bibliography{mybibfile}



\onecolumn
\section*{Appendix}

\section*{Appendix A: Extended Table 2}

We present the extended version of Table 2 of the main text.

\begin{table*}[ht]
\centering
\caption{Performance comparison of GCN models with different width and depth, as well as different placement of the trainable layer.}
\vspace{0.6cm}
\begin{tabular}{|c|c|c|c|c|c|c|}
        \hline
        \multicolumn{7}{|c|}{Accuracy (\%) \& std}\\
        \hline
        \multirow{2}{*}{Dataset} & \multirow{2}{*}{(Width, Depth)} & 
        \multicolumn{5}{|c|}{Position} \\
        \cline{3-7}
        {} & {} & 2 & 4 & 8 & 16 & 32\\
        \hline
        \multirow{6}{*}{Cora} &
        (512,8) & \textbf{79.65 \scriptsize$\pm$ 0.62} & 78.07 \scriptsize$\pm$ 1.31 & 73.49 \scriptsize$\pm$ 2.28 & - & -\\
        \cline{2-7}
        {} & (2048,8) & \textbf{81.85 \scriptsize$\pm$ 0.34} & 81.09 \scriptsize$\pm$ 0.74 & 78.06 \scriptsize$\pm$ 1.12 & - & -\\
        \cline{2-7}
        {} & (8192,8) & 81.25 \scriptsize$\pm$ 0.71 & \textbf{81.47 \scriptsize$\pm$ 0.89} & 81.04 \scriptsize$\pm$ 0.56 & - & -\\
        \cline{2-7}
        {} & (512,32) & 35.29 \scriptsize$\pm$ 12.74 & 36.60 \scriptsize$\pm$ 11.21 & 38.83 \scriptsize$\pm$ 9.03 & \textbf{39.65 \scriptsize$\pm$ 10.99} & 33.55 \scriptsize$\pm$ 12.14\\
        \cline{2-7}
        {} & (2048,32) & 76.90 \scriptsize$\pm$ 2.23 & 77.09 \scriptsize$\pm$ 2.05 & \textbf{77.43 \scriptsize$\pm$ 1.88} & 77.22 \scriptsize$\pm$ 1.95 & 76.99 \scriptsize$\pm$ 2.22\\
        \cline{2-7}
        {} & (8192,32) & 78.87 \scriptsize$\pm$ 0.45 & 78.44 \scriptsize$\pm$ 0.58 & \textbf{79.07 \scriptsize$\pm$ 0.73} & 78.83 \scriptsize$\pm$ 0.43 & 78.97 \scriptsize$\pm$ 0.61\\
        \hline
        \multirow{6}{*}{CiteSeer} & 
        (512, 8) & \textbf{67.58 \scriptsize$\pm$ 1.25} & 62.13 \scriptsize$\pm$ 1.18 & 53.47 \scriptsize$\pm$ 1.85 & - & -\\
        \cline{2-7}
        {} & (2048, 8) & \textbf{71.20 \scriptsize$\pm$ 0.49} & 67.20 \scriptsize$\pm$ 1.18 & 61.14 \scriptsize$\pm$ 1.11 & - & -\\
        \cline{2-7}
        {} & (8192, 8) & \textbf{70.78 \scriptsize$\pm$ 1.45} & 69.52 \scriptsize$\pm$ 1.08 & 66.39 \scriptsize$\pm$ 0.74 & - & -\\
        \cline{2-7}
        {} & (512, 32) & 34.35 \scriptsize$\pm$ 12.25 & 32.40 \scriptsize$\pm$ 9.81 & 29.01 \scriptsize$\pm$ 9.86 & 27.72 \scriptsize$\pm$ 11.78 & \textbf{30.12 \scriptsize$\pm$ 10.94}\\
        \cline{2-7}
        {} & (2048, 32) & 57.36 \scriptsize$\pm$ 6.46 & 55.76 \scriptsize$\pm$ 5.72 & 57.22 \scriptsize$\pm$ 2.87 & 56.72 \scriptsize$\pm$ 3.84 & \textbf{58.69 \scriptsize$\pm$ 3.80}\\
        \cline{2-7}
        {} & (8192, 32) & \textbf{70.18 \scriptsize$\pm$ 0.64} & 68.21 \scriptsize$\pm$ 0.87 & 65.45 \scriptsize$\pm$ 1.06 & 65.45 \scriptsize$\pm$ 0.99 & 65.99 \scriptsize$\pm$ 1.09\\
        \hline
        \multirow{6}{*}{Pubmed} & 
        (512, 8) & \textbf{78.07 \scriptsize$\pm$ 0.85} & 76.99 \scriptsize$\pm$ 0.88 & 72.68 \scriptsize$\pm$ 2.19 & - & -\\
        \cline{2-7}
        {} & (2048, 8) & 77.18 \scriptsize$\pm$ 0.66 & \textbf{77.53 \scriptsize$\pm$ 0.52} & 76.72 \scriptsize$\pm$ 0.66 & - & -\\
        \cline{2-7}
        {} & (8192, 8) & 74.05 \scriptsize$\pm$ 2.67 & 76.20 \scriptsize$\pm$ 1.43 & \textbf{77.75 \scriptsize$\pm$ 0.67} & - & -\\
        \cline{2-7}
        {} & (512, 32) & 66.91 \scriptsize$\pm$ 14.76 & 68.56 \scriptsize$\pm$ 11.22 & \textbf{68.99 \scriptsize$\pm$ 11.79} & 68.32 \scriptsize$\pm$ 13.51 & 65.56 \scriptsize$\pm$ 18.07\\
        \cline{2-7}
        {} & (2048, 32) & \textbf{78.55 \scriptsize$\pm$ 0.40} & 78.22 \scriptsize$\pm$ 0.48 & 78.06 \scriptsize$\pm$ 0.61 & 77.96 \scriptsize$\pm$ 0.93 & 78.24 \scriptsize$\pm$ 0.77\\
        \cline{2-7}
        {} & (8192, 32) & \textbf{78.63 \scriptsize$\pm$ 0.44} & 78.57 \scriptsize$\pm$ 0.51 & 78.53 \scriptsize$\pm$ 0.45 & 78.56 \scriptsize$\pm$ 0.57 & 78.57 \scriptsize$\pm$ 0.45\\
        \hline
        \multirow{6}{*}{Photo} & 
        (512, 8) & \textbf{84.53 \scriptsize$\pm$ 1.32} & 83.54 \scriptsize$\pm$ 0.74 & 66.85 \scriptsize$\pm$ 5.44 & - & -\\
        \cline{2-7}
        {} & (2048, 8) & \textbf{89.68 \scriptsize$\pm$ 0.12} & 89.41 \scriptsize$\pm$ 0.18 & 82.59 \scriptsize$\pm$ 0.36 & - & -\\
        \cline{2-7}
        {} & (8192, 8) & \textbf{90.26 \scriptsize$\pm$ 0.21} & 89.82 \scriptsize$\pm$ 0.22 & 87.73 \scriptsize$\pm$ 0.18 & - & -\\
        \cline{2-7}
        {} & (512, 32) & 22.69 \scriptsize$\pm$ 15.00 & 22.15 \scriptsize$\pm$ 11.70 & 24.42 \scriptsize$\pm$ 15.12 & \textbf{25.68 \scriptsize$\pm$ 15.29} & 21.61 \scriptsize$\pm$ 12.61\\
        \cline{2-7}
        {} & (2048, 32) & 41.12 \scriptsize$\pm$ 17.04 & 44.96 \scriptsize$\pm$ 15.92 & 50.78 \scriptsize$\pm$ 13.71 & \textbf{51.76 \scriptsize$\pm$ 13.70} & 50.19 \scriptsize$\pm$ 13.14\\
        \cline{2-7}
        {} & (8192, 32) & \textbf{72.32 \scriptsize$\pm$ 2.93} & 71.39 \scriptsize$\pm$ 2.04 & 70.58 \scriptsize$\pm$ 1.27 & 70.66 \scriptsize$\pm$ 1.95 & 71.62 \scriptsize$\pm$ 1.80\\
        \hline
        \multirow{6}{*}{Computers} & 
        (512, 8) & \textbf{64.74 \scriptsize$\pm$ 2.14} & 63.81 \scriptsize$\pm$ 1.57 & 59.70 \scriptsize$\pm$ 1.03 & - & -\\
        \cline{2-7}
        {} & (2048, 8) & \textbf{73.10 \scriptsize$\pm$ 0.26} & 72.22 \scriptsize$\pm$ 0.22 & 64.77 \scriptsize$\pm$ 0.71 & - & -\\
        \cline{2-7}
        {} & (4096, 8) & \textbf{74.80 \scriptsize$\pm$ 0.00} & 73.15 \scriptsize$\pm$ 0.00 & 66.52 \scriptsize$\pm$ 0.00 & - & -\\
        \cline{2-7}
        {} & (512, 32) & 8.43 \scriptsize$\pm$ 5.18 & 8.83 \scriptsize$\pm$ 5.24 & \textbf{12.71 \scriptsize$\pm$ 11.89} & 10.82 \scriptsize$\pm$ 5.48 & 8.33 \scriptsize$\pm$ 5.21\\
        \cline{2-7}
        {} & (2048, 32) & 45.74 \scriptsize$\pm$ 19.44 & 48.15 \scriptsize$\pm$ 19.88 & 49.38 \scriptsize$\pm$ 19.14 & 50.85 \scriptsize$\pm$ 17.11 & \textbf{50.97 \scriptsize$\pm$ 17.32}\\
        \cline{2-7}
        {} & (4096, 32) & 65.00 \scriptsize$\pm$ 0.00 & \textbf{67.62 \scriptsize$\pm$ 0.00} & 65.17 \scriptsize$\pm$ 0.00 & 66.28 \scriptsize$\pm$ 0.00 & 64.84 \scriptsize$\pm$ 0.00\\
        \hline
        \multirow{6}{*}{Physics} &
        (512, 8) & \textbf{92.88 \scriptsize$\pm$ 0.14} & 92.68 \scriptsize$\pm$ 0.11 & 90.97 \scriptsize$\pm$ 0.71 & - & -\\
        \cline{2-7}
        {} & (2048, 8) & 92.51 \scriptsize$\pm$ 0.06 & 92.47 \scriptsize$\pm$ 0.12 & \textbf{92.66 \scriptsize$\pm$ 0.17} & - & -\\
        \cline{2-7}
        {} & (8192, 8) & 92.46 \scriptsize$\pm$ 0.06 & 92.41 \scriptsize$\pm$ 0.09 & \textbf{92.65 \scriptsize$\pm$ 0.13} & - & -\\
        \cline{2-7}
        {} & (512, 32) & 39.50 \scriptsize$\pm$ 27.73 & 37.52 \scriptsize$\pm$ 26.90 & \textbf{40.04 \scriptsize$\pm$ 27.52} & 38.29 \scriptsize$\pm$ 26.26 & 33.93 \scriptsize$\pm$ 26.08\\
        \cline{2-7}
        {} & (2048, 32) & 81.54 \scriptsize$\pm$ 24.21 & 80.75 \scriptsize$\pm$ 24.78 & \textbf{86.35 \scriptsize$\pm$ 14.05} & 85.72 \scriptsize$\pm$ 13.94 & 86.09 \scriptsize$\pm$ 13.89\\
        \cline{2-7}
        {} & (8192, 32) & 92.47 \scriptsize$\pm$ 0.18 & \textbf{92.53 \scriptsize$\pm$ 0.16} & 92.44 \scriptsize$\pm$ 0.09 & 92.40 \scriptsize$\pm$ 0.10 & 92.43 \scriptsize$\pm$ 0.12\\
        \hline
        \multirow{6}{*}{CS} &
        (512, 8) & \textbf{87.16 \scriptsize$\pm$ 1.33} & 86.83 \scriptsize$\pm$ 1.39 & 79.82 \scriptsize$\pm$ 1.76 & - & -\\
        \cline{2-7}
        {} & (2048, 8) & 88.34 \scriptsize$\pm$ 0.09 & \textbf{88.45 \scriptsize$\pm$ 0.16} & 88.20 \scriptsize$\pm$ 0.30 & - & -\\
        \cline{2-7}
        {} & (8192, 8) & 87.17 \scriptsize$\pm$ 0.09 & 87.45 \scriptsize$\pm$ 0.08 & \textbf{88.93 \scriptsize$\pm$ 0.12} & - & -\\
        \cline{2-7}
        {} & (512, 32) & 7.17 \scriptsize$\pm$ 5.28 & \textbf{9.04 \scriptsize$\pm$ 7.78} & 5.92 \scriptsize$\pm$ 3.92 & 5.99 \scriptsize$\pm$ 3.75 & 7.30 \scriptsize$\pm$ 3.53\\
        \cline{2-7}
        {} & (2048, 32) & 36.07 \scriptsize$\pm$ 14.23 & 38.16 \scriptsize$\pm$ 10.80 & 41.19 \scriptsize$\pm$ 11.22 & 39.35 \scriptsize$\pm$ 9.42 & \textbf{41.39 \scriptsize$\pm$ 11.59}\\
        \cline{2-7}
        {} & (8192, 32) & \textbf{82.96 \scriptsize$\pm$ 0.92} & 82.32 \scriptsize$\pm$ 1.53 & 82.62 \scriptsize$\pm$ 0.30 & 82.56 \scriptsize$\pm$ 0.36 & 82.37 \scriptsize$\pm$ 0.43\\
        \hline
    \end{tabular}
    \label{tab:pos_inv_ext}
\end{table*}

\end{document}